\pdfoutput=1
\documentclass[11pt]{article}

\usepackage[final]{acl}

\usepackage{times}
\usepackage{latexsym}

\usepackage[T1]{fontenc}
\usepackage[utf8]{inputenc}

\usepackage{microtype}

\usepackage{inconsolata}
\usepackage{amsmath}   % for align* environment
\usepackage{amsfonts}  % for \mathbb
\usepackage{graphicx} % for figures
\usepackage{booktabs}  % for tables
\usepackage{array} % for tables
\usepackage{dcolumn}
\usepackage{lscape} % for landscape

\title{AutoPersuade: \\A Framework for Evaluating and Explaining Persuasive Arguments\thanks{This work was supported by the New Ideas in the Social Sciences fund of Princeton University. The work described in this manuscript is subject to a pending patent application. The replication code is available here: \href{https://github.com/TillRS/AutoPersuade}{https://github.com/TillRS/AutoPersuade}.}}

\author{Till Raphael Saenger \\
  Princeton University\\
  \texttt{saenger@princeton.edu} \\\And
  Musashi Hinck \\
  Intel Labs\thanks{This work was done while MH was a postdoctoral research associate at the Data-Driven Social Science Initiative at Princeton University.}\\
  \texttt{musashi.hinck@intel.com}
  \\\AND
  Justin Grimmer \\
  Stanford University\\
  \texttt{jgrimmer@stanford.edu}\\\And
  Brandon M. Stewart \\
  Princeton University \\
  \texttt{bms4@princeton.edu} 
  }
\begin{document}
\maketitle
\begin{abstract}
We introduce \emph{AutoPersuade}, a three-part framework for constructing persuasive messages. First, we curate a large dataset of arguments with human evaluations. Next, we develop a novel topic model to identify argument features that influence persuasiveness. Finally, we use this model to predict the effectiveness of new arguments and assess the causal impact of different components to provide explanations. We validate AutoPersuade through an experimental study on arguments for veganism, demonstrating its effectiveness with human studies and out-of-sample predictions.
\end{abstract}

\section{Introduction}
Persuasion is a common task in politics, business, government, and our daily lives.  Modern tools---like A/B experiments, surveys, and focus groups---are well-equipped to identify \textit{which} of a pre-existing set of messages is most persuasive, but provide little insight into \textit{what about} them is compelling. Large language models (LLMs) can help to generate new plausibly persuasive messages, but they do not offer causal evidence on whether or how they have succeeded \citep{Gomez-Uribe_Neil_2016_NetflixRecommender, deVaus2013surveys, Morgan1996_FocusGroups, PalmerStirling_2023_LLMs, rescala2024can}. 

\begin{figure*}[ht]
    \makebox[\textwidth][c]{
        \includegraphics[width=1.05\textwidth]{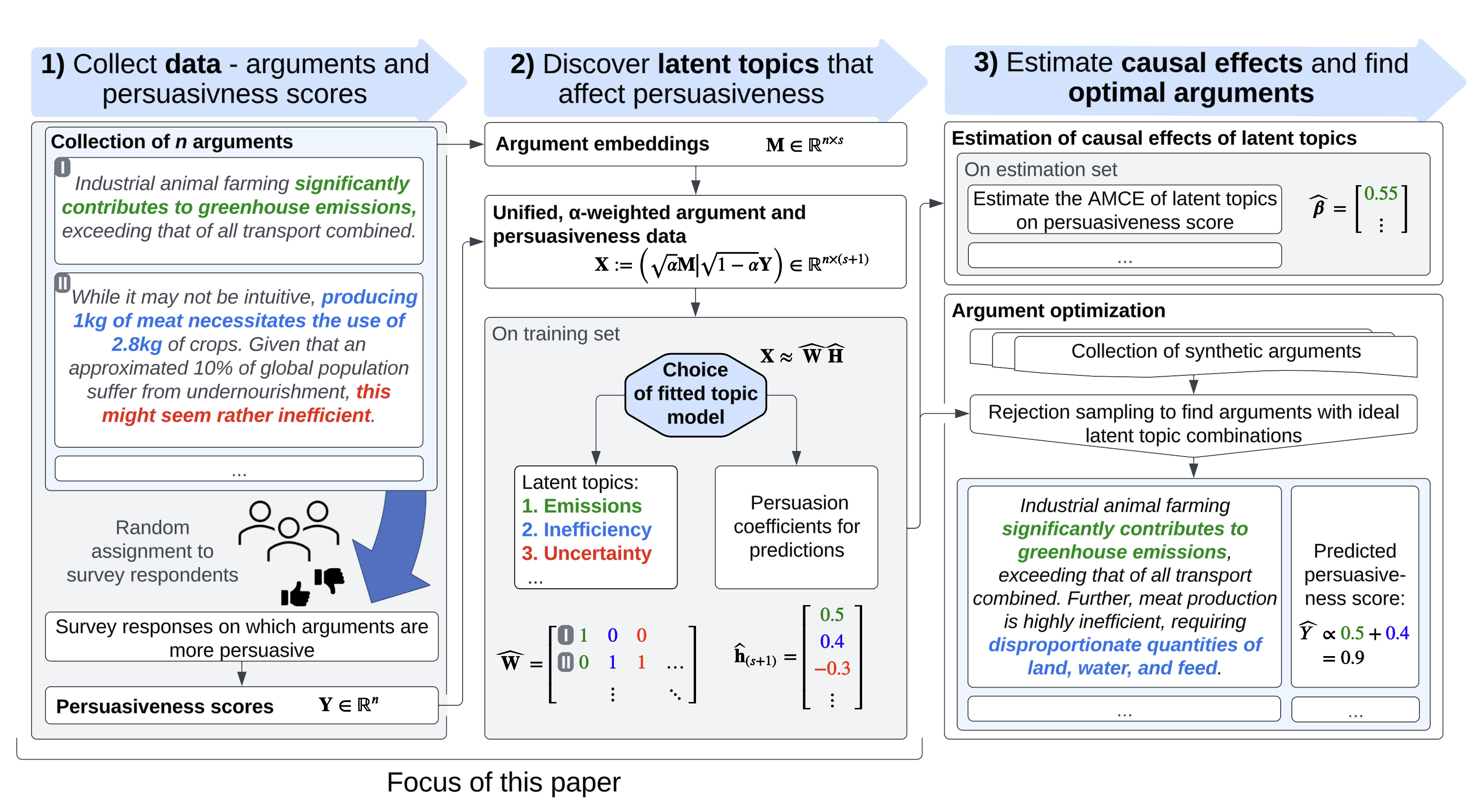}%
    }
    \caption{The \emph{AutoPersuade} workflow. After generating a collection of arguments, we collect reactions from respondents. Using these reactions and arguments, we fit the SUN topic model that discovers latent topics that simultaneously describe the documents and their persuasiveness.  We then use the output from the SUN model to estimate causal effects from the existing sample and to predict the persuasiveness of new arguments. While our validation studies confirm the causal estimates of our case study, the current approach using average marginal component effects of topics appears less well-suited to finding the best argument. Improving the identification of optimal arguments in the last step might be the subject of future studies.
    }
    \label{fig:Workflow_arguments}
\end{figure*}

In this paper, we introduce a new workflow for identifying the topical components of an argument that are persuasive, \emph{AutoPersuade}.  Our framework assists with each step of the persuasion task.  Our three-step workflow is shown in Figure \ref{fig:Workflow_arguments}. We demonstrate this workflow in a novel study of pro-veganism persuasion.

First, we gather persuasive arguments and responses to those arguments. These arguments can be from various sources, like social media, LLMs, or manual generation. With such a collection of arguments, we discuss our exemplary case study to evaluate arguments for veganism. Using a forced-choice design with arguments randomly assigned to respondents, this case study allows us to assess which arguments survey respondents report as more persuasive. While self-reported assessments of persuasion may not correspond to actual changes in behavior \citep{coppock2023persuasion}, our new framework is well suited for alternative settings that use behavioral measures of persuasion, which might be explored in future studies.

Second, with arguments and responses in hand, we introduce a new model to extract the latent features that cause arguments to be more or less persuasive. We call this model the \underline{SU}pervised semi-\underline{N}on-negative (SUN) topic model. The SUN topic model uses an embedding representation of the arguments and builds upon matrix factorization methods to extract latent features that characterize arguments and affect responses \citep{fong2016sIBP, egami_2022_CausalInf_SienceAdv, feder-etal-2022-causal}. Because our model provides interpretable output, we show how it can be used to provide easy-to-understand and actionable insights into why particular messages are (or are not) persuasive.   

Third, using the output from the SUN topic model, we compute causal effects from varying the content of our arguments and assess the persuasiveness of future arguments. We target causal effects that determine how the average persuasiveness of an argument changes as the prevalence of a latent feature changes. 
This allows us to predict how persuasive a new argument would be if deployed to the same population. However, as we demonstrate through our experiments, our current method is better at answering questions about what is (and is not) persuasive than finding the optimal persuasive argument.  

Our quantity of interest is in the \textit{content} of arguments that are persuasive in particular settings rather than the \textit{rhetorical strategies} that are effective across settings. In this sense we are pursuing a different goal than, e.g., the pioneering work of \citet{Tan2016_CMV_Reddit} which considers linguistic features that are persuasive in online discourse.

\section{The AutoPersuade Workflow}
\label{sec:Methods_2}
In this section, we describe the three-step AutoPersuade workflow (summarized in Figure \ref{fig:Workflow_arguments}) in detail. 

\subsection{Step 1: Collect Data}

Arguments in this paper are a collection $\mathcal{D}$ of documents from which $n$ distinct samples are shown to respondents from a well-defined survey population.  Arguments elicit a reaction, which we collect in $\mathbf{Y}$. Unlike A/B tests, our workflow accommodates, and indeed benefits from, a large number of potential arguments  \citep{fong2023causal}.  Ideally, the collected responses will reflect the respondent's behavior. That said, we can also use proxies for that behavior including respondent's self-reported evaluations of statements.

\subsection{Step 2: Discover Topics}
\label{sec:TopicModel_2.1}

The SUN topic model takes the arguments and responses and extracts the latent features underlying the arguments that are driving the responses.  Once estimated, this model enables us to identify why certain arguments are more or less persuasive. 
  
To apply the SUN model we represent each argument as an element of $\mathbb{R}^{s}$ using a document embedding, where $s$ depends on the dimensionality of the specific embedding used. After representing each of our arguments as an embedding, our collection of arguments is $\mathbf{M} \in \mathbb{R}^{(n\times s)}$.

Using this representation the SUN topic model builds on prior work on unsupervised topic models \citep{Lee1999_NMF, Blei07, fong2016sIBP}. These models typically use a bag-of-words representation that is non-negative. Because our data representation (the embeddings) includes negative values, we utilize a semi-non-negative matrix factorization that allows negative values in the data representation but still guarantees nonnegative topics. 
While, as with all topic models, the feature loadings produced by the SUN topic model are not intrinsically interpretable, they enable us to find lower-dimensional representations that can be interpretable when finding a good model fit and assigning appropriate topic labels.  
Particularly, we find that the non-negativity constraint on the loadings promotes interpretable latent factors.

To avoid issues with the definition of causal effects, we do not constrain the prevalence of the causal effects to sum to 1 \citep{fong2016sIBP} as would occur in classic LDA topic model variants \citep{blei2003latent}. Unlike prior work \citep{fong2016sIBP, fong2023causal}, we do not discretize the loadings as present or absent and instead allow for topics to have a non-negative prevalence across documents. We place no constraints on the effect that the presence of a latent topic might have on the responses $\mathbf Y$.

\paragraph{SUN Model Setup:} The SUN topic model discovers and estimates latent features that simultaneously explain differences in the arguments and in the responses to the arguments.\footnote{Topic model implementation is available here: \href{https://github.com/TillRS/SUN_TopicModel}{https://github.com/TillRS/SUN\_TopicModel}.} 
We build on \citet{Ding2008_SemiNMF} to discover the latent topics for the embedding representation, $\mathbf{M}$, of the arguments, such that
\begin{align}
\label{eq:SNMF_approximation}
    \mathbf{M} &\approx \mathbf{WB}  \ \text{where } \mathbf{M} \in \mathbb{R}^{n \times s},  \mathbf{W} \in \mathbb{R}_+^{n \times J}, \\ 
    \notag & \quad \quad \quad \quad \quad \text{ and }  \mathbf{B} \in \mathbb{R}^{J \times s}
\end{align}
where $J \in \mathbb{N}^{+}$ is a user-set parameter that determines the number of topics.
Here, each row of $\mathbf{W}$ captures the presence, or loadings, of each latent topic for a given document while $\mathbf{B}$ is the mapping between these latent topics and the embedding space.

Because we want our latent topics to explain the responses $\mathbf Y$, we also consider 
\begin{align}
\label{eq:response_model}
    \mathbf{Y} & \approx \mathbf{W}\boldsymbol{\gamma} \quad \text{ where } \mathbf{Y} \in \mathbb{R}^n, \boldsymbol{\gamma} \in \mathbb{R}^J.
\end{align}
Here, $\boldsymbol{\gamma}$ captures the relationship between the latent variables and the responses. We refer to $\boldsymbol{\gamma}$ as the \textit{persuasion coefficients}.  These coefficients will be an important determinant of the causal effects we estimate later, but may not justify a causal interpretation when analyzed directly from the model.

\paragraph{Defining the SUN Topic Model Loss Function:}
The total loss function for the SUN topic model is a convex combination of a loss function for the model to explain the latent features in the arguments and a loss function for the latent features that best explain the responses.  

We define the first component of our loss function corresponding to our approximation of $\mathbf{M}$ in \eqref{eq:SNMF_approximation} as 
\begin{equation*}
    \mathcal{L}_A = \frac{1}{2}\left\|\mathbf{M}-\mathbf{W} \mathbf{B} \right\|_F^2 
\end{equation*}
where $\|\cdot \|_F$ denotes the Frobenius norm.  Next, we define the loss for our approximation of the responses $\mathbf{Y}$ in \eqref{eq:response_model} as
\begin{equation*}
    \mathcal{L}_R = \frac{1}{2}\left\|\mathbf{Y}-\mathbf{W} \boldsymbol{\gamma}   \right\|_2^2.
\end{equation*}

We combine the two loss functions to define the total loss function as 
\begin{equation*}
    \mathcal{L}  = \alpha \mathcal{L}_A + (1-\alpha)\mathcal{L}_R
\end{equation*}
where $\alpha \in (0,1)$ is a parameter that controls the share of weight placed on the argument loss function $\mathcal{L}_A$ or the response loss function $\mathcal{L}_R$. As $\alpha$ goes to one, the latent topics are increasingly focused on explaining the content of the arguments in the embedding space, $\mathbf{M}$.  As $\alpha$ goes to zero, our latent topics are increasingly focused on only explaining the responses, $\mathbf{Y}$. The $\alpha$ parameter enables us to discover latent topic combinations that balance explaining variation in the documents and in the responses.   

Further, we can perform simple algebraic manipulation, as included in the Appendix \ref{sec:Appendix_Total_Loss}, to rewrite the total loss function as
\begin{align}
\label{eq:total_loss}
    \notag \mathcal{L} & = \alpha \frac{1}{2}\left\|\mathbf{M}-\mathbf{W} \mathbf{B}\right\|_F^2  + (1-\alpha) \frac{1}{2}\left\|\mathbf{Y}-\mathbf{W} \boldsymbol{\gamma}  \right\|_2^2 \\ 
                    & = \frac{1}{2} \left\| \mathbf{X}  - \mathbf{W} \mathbf{H} \right\|_F^2 
\end{align}
where $\mathbf{X}: = \left(\sqrt{\alpha}\mathbf{M} \big| \sqrt{1-\alpha} \mathbf{Y} \right)$ and $\mathbf{H} : = \left( \sqrt{\alpha}\mathbf{B} \big| \sqrt{1-\alpha} \boldsymbol{\gamma} \right)$. In other words, algebraic manipulation reduces the supervised problem to the well-studied problem of semi-nonnegative matrix factorization. 

\paragraph{Estimating the SUN Topic Model:}
Minimizing \eqref{eq:total_loss} is a non-convex optimization problem, but we can use the closed form updating steps of \citet{Ding2008_SemiNMF}. We detail those straightforward updating steps in section \ref{sec:appendix_SNMFupdates} of the Appendix. The estimation routine proceeds with iterative updates of $\mathbf{W}$ and $\mathbf{H}$ for a fixed number of steps (early stopping), or until we reach convergence. The results of this estimation procedure are the estimated topic loadings $\widehat{\mathbf{W}}$ and the estimated relationship $\widehat{\mathbf{H}}$ between the topic loadings and the unified, scaled input data $\mathbf{X}$. Importantly, the column $\widehat{\mathbf{h}}_{(s+1)} = \sqrt{1-\alpha} \widehat{\boldsymbol{\gamma}}$ captures the estimate for the scaled persuasion coefficients.

As with other topic models, each model fit only corresponds to a local minimum of the non-convex total loss function \eqref{eq:total_loss}. This means that we cannot rely on optimization alone to choose a model for analysis.  Instead, we evaluate models using both numerical and qualitative information.  
Quantitatively, we run 10-fold cross-validation across different choices for $J$ and $\alpha$, where we perform out-of-sample prediction using the procedure described in Section \ref{s:insight}. We use numerical information about the out-of-sample predictive accuracy of the model for predicting the responses because this will correspond directly to the task of forecasting an argument's persuasiveness. We also discuss the average topic coherence metric of \citet{mimno-etal-2011-optimizing} for the chosen topic model fit in our case study.
Qualitatively, we manually inspect the coherence and exclusivity of the latent topics for different model fits \citep{roberts2016model}. To perform this evaluation, we inspect elements in the training set with very high, mid, and low loadings of a given latent topic and try to identify common themes and features. If we view these topics as coherent, mutually distinct, and plausible, we deem it to be a good fit in terms of latent topics. 

\subsection{Step 3: Estimate Causal Effects} \label{s:insight}    

After fitting the SUN topic model, we use it to estimate the causal effect of changing a topic's presence in the documents and to forecast the persuasiveness of new arguments. Both of these tasks require inferring topics for a new argument not included in the original model fit. When inferring topic loadings of a new document $t \in \mathcal{D}$, we do not include its response $\mathbf{Y}_t $ as model input, as we are trying to estimate the relationship to the response or to predict it. 
Hence, we are trying to learn the topic loadings solely based on the embedding $\mathbf{M}_t$ after standardizing and scaling it following the training data scaling. 
This corresponds to using $\sqrt{\alpha}\mathbf{M}_t$ and minimizing $\mathcal{L}_A$ to infer the latent topic loadings $\widehat{\mathbf{W}}_t$ while holding $\sqrt{\alpha} \widehat{\mathbf{B}}$ fixed. We leave $\widehat{\mathbf{B}}$ unchanged to preserve the previously discovered latent topics and focus on learning the presence of these topics in the new argument $t$. 
Note that we can extract $\sqrt{\alpha} \widehat{\mathbf{B}}$ from the previously derived $\widehat{\mathbf{H}}$ and minimize $\mathcal{L}_A$ by initializing and updating $\widehat{\mathbf{W}}_t$ while holding $\widehat{\mathbf{B}}$ constant\footnote{We derive the original model fit, $\widehat{\mathbf{W}}$ and $ \widehat{\mathbf{H}}$, using 100 updating steps. Matching this, we originally used 100 updating steps when inferring topic loadings for new documents to create the presented results. 
However, the convex topic inference problem does not consistently converge after 100 steps. Hence, we implemented an alternative approach using CVXPY \citep{diamond2016cvxpy} to infer topic loadings. While our main results remain largely unchanged, we report the results of this alternative method in Appendix \ref{Appendix:New_topic_inf} and briefly discuss some benefits of early stopping and running to convergence, respectively.}.

To avoid theoretical issues that arise when fitting a latent model and estimating the causal effect of those same latent topics, we use an estimation set of arguments that had not been previously used for model fitting \citep{egami_2022_CausalInf_SienceAdv}.  

More formally, let $\mathbf{M}_E$ and $\mathbf{Y}_E$ denote the document embeddings and responses of the estimation set. We infer the latent topics $\mathbf{W}_E$ as described above. Then we can fit the regression 
\begin{equation*}
    \mathbf{Y}_E = \widehat{\mathbf{W}}_E \boldsymbol{\beta} + \mathbf{e}
\end{equation*}
where $\boldsymbol{\beta}$ is the estimand of the instantaneous effect of the latent topics under the assumption that the topics are independent of $\mathbf{e}$, the error term. We focus on the effects of topics estimated using a linear model, but the AutoPersuade workflow is modular: it works with any model that infers the causal effects of underlying latent features of the arguments.  %This means that our estimation of causal effects is not limited to average treatment effects but allows for estimating any other causal relationship between the latent topics and responses that might be of interest in a given application. 

As a second application, we use the output of the SUN topic model to forecast the response to new arguments.  This is particularly powerful because it enables us to quickly identify arguments that our evidence suggests will be persuasive if deployed. These can be generated effectively using LLMs. As we will show, this works well \textit{on average} but not for optimizing the \textit{most persuasive} arguments.  This in turn makes the technique best suited to answering explanatory questions.

%Assessing the persuasiveness of new arguments requires that we have new arguments in hand. Researchers and substantive experts can use the information from the SUN topic model fit to generate new arguments composed of the most persuasive topics. Alternatively, we carefully design prompts to an LLM to induce it to construct arguments focused on the most persuasive features, similar to our human experts.  %Ideally, our prompts will induce the LLM to construct arguments that emphasize the most persuasive features of the arguments.  

%We then use a variety of procedures to identify the best argument to deploy in the next round of study.  Regardless of how an argument is constructed---whether from a creative ad agency or an LLM--- we can assess its expected persuasion score.  Our approach also enables us to screen arguments that might backfire.  It is straightforward to implement a rejection sampling approach that removes synthetic arguments that fall below some minimum threshold the researcher determines.  

\subsection{Designing Baselines}
The challenge in evaluating and comparing our method to strong baselines is that it provides two distinct outputs: Predictions of the persuasiveness of held-out messages and, more importantly, estimates of interpretable components of the message and their causal effects. Because few methods attempt to do both tasks, we assess performance with respect to baselines that maximize each component separately.

The strongest baseline for predictive performance sacrifices interpretability to maximize predictivity directly. We compare our topic model results to classic supervised baseline methods based on the same document embeddings (Lasso, Gradient Boosting, and Random Forest) which can asymptotically approximate more complex functions of the embeddings than our topic models to predict persuasiveness. Although these are older methods, they work well with relatively small datasets and they provide a direct measure of what we are giving up to obtain interpretable estimates.

It is difficult to assess how well we extract interpretable components and their causal effects (since there is neither a clear measure of interpretability nor a way to observe a causal effect). 
Here we design a series of human validation studies that compare our approach to the strong baseline of asking large language models to improve on existing persuasive arguments and construct new persuasive arguments.

\section{Case Study: Pro-Veganism Arguments}
\label{sec:Vegansim}
We use a survey experiment to collect respondents' reactions to pro-veganism arguments. We ask each respondent to compare two messages and to choose the one they find more persuasive. We then summarize these pairwise contests using a Bradley-Terry model and use this as our response variable \citep{bradley_terry1952, Newman2023_BT_Score}. Finally, we use the arguments and the summarized performances to fit the SUN topic model and evaluate the persuasiveness of new potential arguments.

In particular, we start out with an original collection of arguments and responses followed by three validation studies to validate our estimates and explore the predictability of argument performances.
Validation Studies 1 \& 2 focused on generating and comparing synthesized and modified arguments with high persuasiveness scores to test whether we can improve on the best-performing arguments of the original argument collection. 
However, we find that such filtering based on the average marginal component effect of latent topics does not reliably improve arguments within the tail of best-performing arguments.
Validation Study 3, on the other hand, validates our estimate of the average marginal component effect on argument persuasiveness by intervening on arguments across the entire distribution of topic loadings, not just a tail. 

\subsection{Curating Pro-Vegan Messages}
As a first step, we created the original argument collection used in our survey evaluation.
We began by curating a set of 93 root arguments, primarily sourced from longer essays on animal rights from advocacy websites. These arguments were distilled into shorter versions (approximately 280 characters) using GPT-4 \citep{OpenAI_GPT4_2023}. This distillation process involved summarizing each argument into short statements of circa 160 characters and then expanding these summaries back into 280-character versions.
Each of the 93 original arguments was summarized three times, and GPT-4 was used to generate five "more persuasive" versions of the first two summaries and three "less persuasive" versions of the third summary. We prompted for less and more persuasive versions to validate if our survey-based results aligned with these instructions on how to paraphrase the arguments.
Overall, this yielded a total of 1209 arguments based on the original 93 arguments. 

Additionally, 100 arguments were generated solely by GPT-4 without any prior argument collection. These 100 arguments were created by prompting GPT-4 to produce ten distinct pro-veganism arguments, which were then summarized and expanded as described above. Here, we did not prompt for any of them to be "less persuasive".
The complete original argument collection thus consists of 1309 pro-veganism arguments.
This collection of arguments, including original sources and intermediary summaries as well as the arguments generated for later validation studies, are provided in the Supplementary Materials.

\subsection{Collecting and Summarizing Responses}
\label{sec:Veganism_Collecting_Responses}
After curating the original collection of arguments, we divided it into a training set, comprising $2/3$ of the arguments, and an estimation set, comprising the remaining $1/3$. This division is stratified based on the 93 underlying root arguments, ensuring our training and test sets are well-balanced.  

We then deployed a survey on Amazon's Mechanical Turk (MTurk) platform to evaluate the arguments. 
In this survey, each respondent was shown a pair of arguments about veganism and then asked to select the more persuasive argument. 
The argument comparisons were fully randomized, both the pairwise comparisons and the order in which the arguments appeared (displayed side by side).
In total, each respondent evaluated five pairwise comparisons.
We included the full survey questionnaire in the Supplementary Materials and additional details on survey results in the Appendix. 

We compiled the results of 1,036 sessions with five pairwise comparisons of arguments each. This results in 5,180 total evaluations of pairs of arguments. To increase our sample size, we allowed respondents to complete multiple sessions.
To summarize an argument's performance across the comparisons, we fit a Bradley-Terry (BT) model \citep{bradley_terry1952, Newman2023_BT_Score} on the training set arguments. The BT model ranks arguments in terms of their likelihood to win a pairwise contest, and we use this summary as our response variable for the SUN topic model.  We infer a test set document's response variable by applying the Bradley-Terry model to its performance in pairwise contests while keeping the training set arguments' scores fixed. 

Note that we consider this argument evaluation to be a survey and not an annotation task.  In that sense, disagreement is expected and appropriate, as different people are persuaded by different arguments. We measure argument persuasiveness based on average response within this population.

Of the 5180 comparisons on the original argument collection, 49.9\% of the arguments on the right-hand side won the pairwise comparisons, suggesting no ordering effects.
Further, while we found some evidence that argument length affects performance positively, it is uncorrelated with the latent topics discovered in our argument collections and thus does not meaningfully affect our estimated effects. The estimates presented in section \ref{sec:Veganism_Estimating_AMCE} are derived while controlling for argument length.

Lastly, the arguments of the original collection, prompted to be 'more convincing' achieved an average Bradley-Terry score of 1.03 while the 'less convincing' arguments averaged 0.96. As these two groups are stratified across the underlying root arguments, this indicates that the GPT-prompting and our argument performance evaluation achieve and identify the desired effects.

\subsection{Fitting the SUN topic model}
\label{sec:Veganism_Topcis}
We represent the arguments in our collection in an embedding space using OpenAI's "text-embedding-ada-002" model \citep{OpenAI_2022embeddings}.
However, we tested and found similar predictive performance using alternative embedding models, as outlined in the Appendix.
Equipped with this data, we fit several instances of the SUN topic model to arrive at a final model choice. We ran 10-fold cross-validation on the training data across different topic numbers $J$ and $\alpha$ values to explore how predictive performance and topic coherence and exclusivity changed under different parameter choices and local minima.

\begin{figure}[ht]
    \centering
    \includegraphics[width=0.5\textwidth]{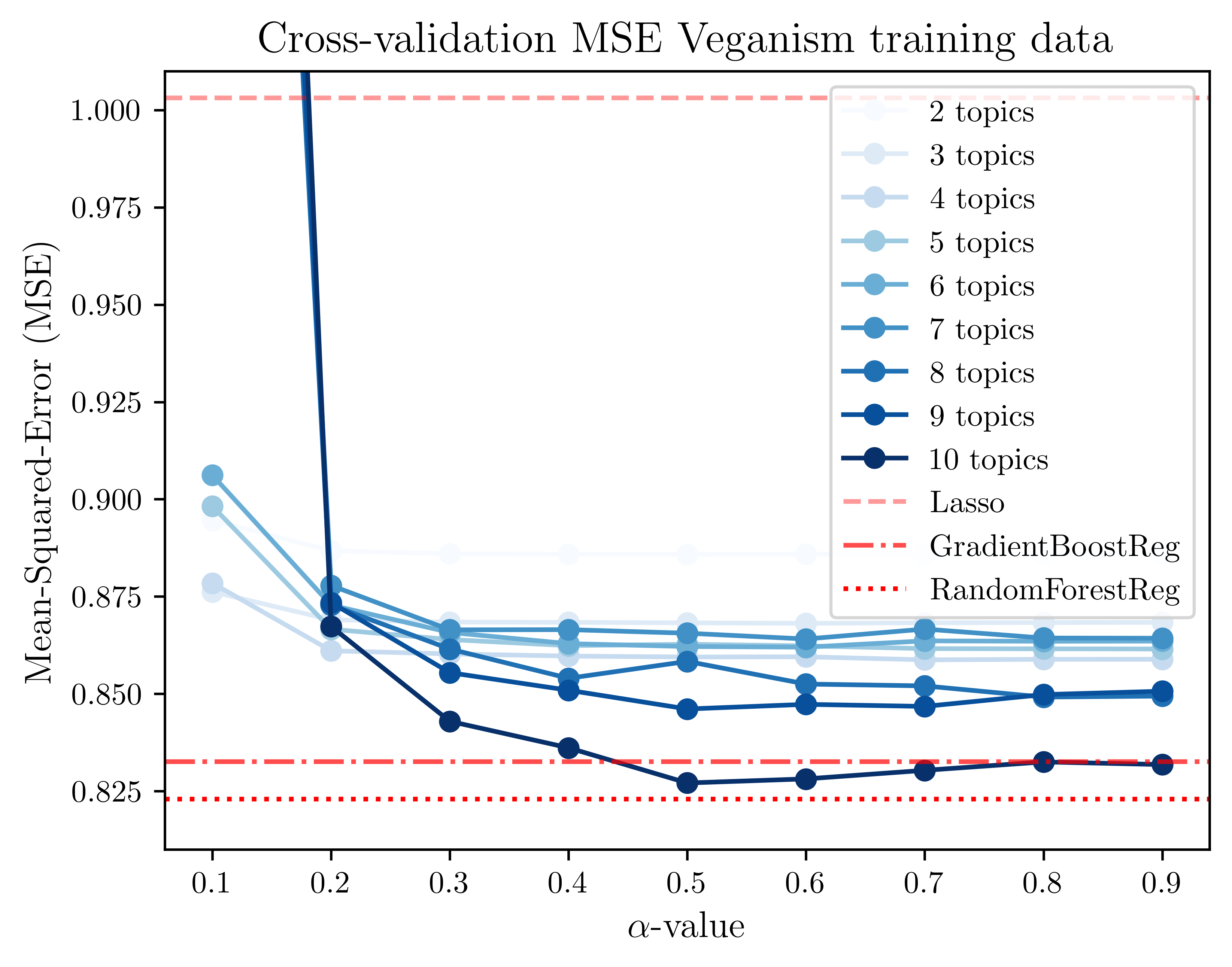}
    \caption{Out-of-sample predictive accuracy of SUN topic model for different parameter choices, as well as benchmark models on the training data. Results were calculated using 10-fold cross-validation.}
    \label{fig:Veganism_CV}
\end{figure}

Using the output from the SUN topic model we found that 10 topics and $\alpha = 0.5$ performed well both relative to other hyperparameter choices and benchmark models as shown in Figure \ref{fig:Veganism_CV} (recalling that even matching a classic supervised baseline should be challenging because they are not constrained to working with interpretable components). We then fit multiple models with these parameter choices, but different random initializations. For this model selection step, we used 80\% of our training data and checked the predictive performance on the remaining 20\% to prevent us from overfitting.

Table \ref{tab:Veganism_Topics} provides our human-assigned labels for our SUN topics. The model achieves an in-sample MSE of $0.82$ and an out-of-sample MSE of $0.7$ on our training data set. The higher in-sample MSE is likely driven by outliers. 

\begin{table}
    \centering
    \small
    \begin{tabular}{c p{6.5cm}}
        \hline
         & \textbf{Latent Topics} \\
        \hline
        (1) & Uncertainty and generalizations \\
        (2) & Inefficient use of resources \\
        (3) & Exploitation, suffering, and compassion\\
        (4) & Morals, ethics, and justifications\\
        (5) & Treatment of cows and chickens \\
        (6) & Individual contributions and responsibility \\
        (7) & Animal rights and speciesism \\
        (8) & Health benefits \\
        (9) & Addressing criticism and fallacies \\
        (10) & Climate change and sustainability \\
        \hline
    \end{tabular}
    \caption{Labels for the discovered latent topics of the arguments for veganism.}
    \label{tab:Veganism_Topics}
\end{table}

The selection of an optimal model fit with the chosen $\alpha = 0.5$ and $J = 10$ hyperparameters primarily relies on inspecting documents with high topic loadings and manually identifying themes. However, we also explored numerical approaches for model fit selection. Specifically, we examined the topic coherence metric proposed by \citet{mimno-etal-2011-optimizing}, which measures how frequently the most common words of a given topic co-occur. This metric, based on word frequencies, is typically used with bag-of-words representations.

For our embeddings-based topic model, we identified the most frequent words per topic by extracting the 25 documents with the highest loadings for each topic and computing TF-IDF scores \citep{salton1988tfidf}. We calculated the topic coherence metric using the 5 words with the highest TF-IDF scores for each topic.

This approach highlights the trade-off between finding the most coherent topics and identifying topics that best explain variations in the response variable. A high coherence score and a low out-of-sample prediction MSE characterize good performance. While this specific topic coherence metric is just one of many possible ways to evaluate topics selected by a topic model, and we generally emphasize the importance of the manual inspection process, Figure \ref{fig:Veganism_Topic_Coherence} in the Appendix demonstrates that our selected model fit presents a well-performing balance of these two metrics relative to other random model fits.

\subsection{Estimating Causal Effects}
\label{sec:Veganism_Estimating_AMCE}

Using our model fit, we inferred the latent topic loadings of the arguments in our estimation set. Combining these with the inferred Bradley-Terry scores, we estimated the average marginal component effect of the topics on a document's relative performance in the pairwise contests \citep{hainmueller2014causal, fong2023causal}.

\begin{figure}[ht]
    \centering
    \includegraphics[width=0.5\textwidth]{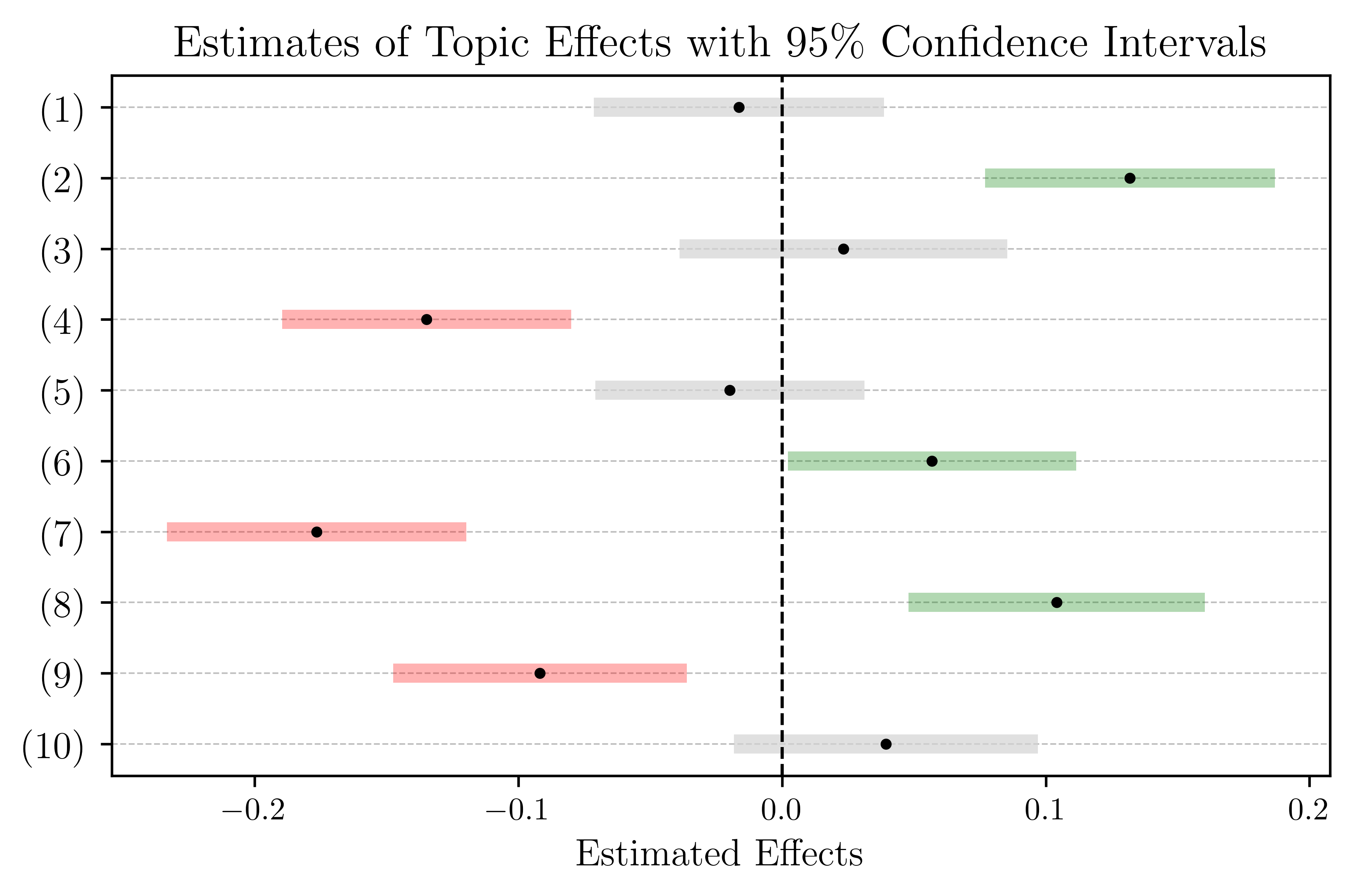}
    \caption{Estimated effects of discovered latent topics on the persuasiveness score of arguments for veganism. Refer to Table \ref{tab:Veganism_Estimation} in the Appendix for more details.}
    \label{fig:Veganism_Effects}
\end{figure}

Figure \ref{fig:Veganism_Effects} shows the estimates and their respective confidence intervals when controlling for argument length. The latent topics associated with significant positive effects are describing the inefficient use of resources for animal products (2), highlighting the importance and impact of individual consumer choices (6), and health benefits (8). On the other hand, discussing morals and ethical justifications for meat consumption (4), animal rights and so-called speciesism \citep{singer2009speciesism} (7), and addressing criticism of veganism and its supposed fallacies (9) are associated with negative effects on our persuasiveness score. The effects of the other discovered latent topics are estimated to be close to zero.

\subsection{Validation Studies}
\label{sec:Validation_Studies}
The challenge of evaluating causal effects is that there is no perfect benchmark because we can never observe a causal effect \citep{feder-etal-2022-causal}. An observable implication of learning what we hope to learn is that we can manipulate arguments to make them more persuasive.  We test that implication here across three human-validation studies.

\subsubsection{Validation Study 1: Generating New Persuasive Arguments}
In our first validation study, we evaluated the persuasiveness of newly synthesized and modified arguments generated from the 45 best-performing arguments in our original collection of 1309, based on Bradley-Terry scores from the initial survey experiment. We identified 20 of these top arguments as \emph{proto-arguments} and used them to create new arguments via two approaches: synthesis and stronger emphasis.

\paragraph{Synthesis Arguments:} We prompted GPT-4 to combine pairs of proto-arguments, generating $(20 \times 19) \times 2 = 760$ synthesis arguments.

\paragraph{Stronger Emphasis Arguments:} We prompted GPT-4 to rewrite each proto-argument with increased emphasis on its primary latent topic, generating three new versions for each proto-argument, resulting in $3 \times 20 = 60$ stronger emphasis arguments.
\\
\\
For both approaches, we filtered the new arguments using rejection sampling to check whether they achieved the desired latent topic loadings and had higher predicted persuasiveness scores than the underlying proto-arguments.
From these filtered arguments, we selected 30 distinct synthesis arguments and 15 distinct stronger emphasis arguments based on their predicted persuasiveness scores. We also generated 10 arguments by prompting GPT-4 to produce its “most persuasive” pro-veganism argument, resulting in a final set of 100 arguments: 45 best-performing original arguments [Original], 30 synthesis arguments [Argument Synthesis], 15 stronger emphasis arguments [Stronger Emphasis], and 10 best  GPT arguments [GPT-best]. 

Using the same pairwise comparison strategy of the original setup, we evaluated the performance of these arguments against each other. 
We collected responses to 990 pairwise comparisons among the arguments from  198 unique MTurk respondents.
Table \ref{tab:Veganism_Validation_WinRates} shows that the synthetic arguments won 54\% of the time, outperforming the stronger emphasis arguments (51.8\%), the GPT-best arguments (51.1\%), and the original arguments (45.4\%) as shown in Table \ref{tab:Veganism_Validation_WinRates}.   
The confidence interval for the difference in win rates between Argument Synthesis and Original ($\text{SY} - \text{OG}$) is $[1.5, 15.7]$, and for the difference between Argument Synthesis and GPT-best ($\text{SY} - \text{GPT}$), it is $[-5.5, 12.1]$.
In other words, synthesizing arguments by combining the best properties determined by our workflow beats the best of the original arguments and appears to outperform the baseline of asking GPT-4 to make its best argument.

\begin{table}
    \centering
    \small
    \begin{tabular}{lrr}
    \toprule
     & Win Rate (\%) & 95\% CI \\
     \midrule
    \multicolumn{3}{l}{\textbf{Validation Study 1}} \\
    Argument Synthesis (SY) &\textbf{54.0} & [49.9, 58.3] \\
    Original (OG) & 45.4 & [41.4, 49.1] \\
    GPT-best (GPT) & 51.1 & [44.4, 57.1] \\
    Stronger Emphasis (SE) & \underline{51.8} & [46.3, 56.8] \\
    \midrule
    \multicolumn{3}{l}{\textbf{Validation Study 2}} \\
    Argument Synthesis (SY) & \underline{48.5} & [44.3, 53.1] \\
    Original (OG) & \textbf{52.0} & [47.4, 56.3] \\
    Stronger Emphasis (SE) & 45.9 & [35.8, 56.4] \\
    \midrule
    \multicolumn{3}{l}{\textbf{Validation Study 3}} \\
    Increased Topic (2) Args. & \textbf{69.7} & [65.3, 74.1] \\
    Decreased Topic (2) Args.  & 31.7 & [22.7, 41.4] \\
    \bottomrule
    \end{tabular}
    \caption{Results of Validation Study 1, 2, and 3. For each study, win rates are calculated as the share of comparisons with arguments of other origins that are won. Confidence intervals are calculated based on 500 bootstraps of the individual comparison outcomes. The highest score for each study is bolded, the second-highest score is underlined.}
    \label{tab:Veganism_Validation_WinRates}
\end{table}

\subsubsection{Validation Study 2: Limitations of Argument Optimization}
Building on Validation Study 1, Validation Study 2 aimed to assess the limitations of our argument optimization approach by directly comparing the synthesized arguments to their respective proto-arguments. No new arguments were generated for this study; instead, we focused on the 30 synthesis and 15 stronger emphasis arguments and compared them against the original arguments from which they were derived. 
While the order of the compared arguments (left vs. right) was still randomized, the pairings were now fixed, as each new argument was only compared against its corresponding proto-arguments.

We collected responses from 102 unique MTurk respondents, totaling 510 random pairwise comparisons.
The results summarized in Table \ref{tab:Veganism_Validation_WinRates} indicate that the synthetic arguments did not consistently outperform their original counterparts. This suggests that while our method successfully identified persuasive topics, enhancing these topics at the higher end of the distribution did not result in significantly more persuasive arguments. The findings suggest an open challenge for maximizing argument persuasiveness.

\subsubsection{Validation 3: Evaluating the Average Effects in the Population}
The causal effects we estimate in our design are the effect of interventions defined over the full population of documents, the Average Marginal Component Effect (AMCE).  To generate a validation based on this estimand, we focus on a single topic: Topic (2) \textit{inefficient use of resources}. 
For this validation study, we intervened on randomly selected arguments from the original set of 1309 arguments.
We first selected 100 random arguments with a high topic (2) loading (greater than 2.0, or the 87th percentile) and prompted GPT-4 to rewrite them to decrease the presence of topic (2). Next, we selected 200 arguments with a low topic (2) loading (less than 2.0) and prompted GPT-4 to rewrite these arguments to increase the presence of topic (2). From these alterations, we curated a collection of 80 arguments with increased topic (2) presence and 20 with decreased presence, alongside the original 100 arguments, forming the 200 arguments used for Validation Study 3.

We collected responses from 100 unique MTurk respondents, totaling 500 pairwise comparisons. Again, we only allowed for comparisons of a given new, altered argument against its underlying, original argument. 
Table \ref{tab:Veganism_Validation_WinRates} shows that manipulations of the documents behave as expected when averaged across the entire distribution.
\\
\\
The interventions used in the validation studies provide a face-validity check that our labels correspond well to our learned latent topics by demonstrating that adjusting arguments to emphasize a specific theme increased or decreased the associated topic loadings. A more detailed discussion of the rejection sampling process used in the validation studies is provided in the Appendix.

Together, our validation studies strongly suggest that the intervention mechanism informed by the AMCE has the expected effect, which allows us to validate topic labels and AMCE estimates on new samples. 
However, improving the best-performing arguments of our sample based on these AMCE insights suggests that the effect on the margin differs from the effect in the tail of the distribution. While this is great for explanation, the method is less well-suited to identifying the best argument. 

\section{Related Work}
\label{sec:RelatedWork}
The AutoPersuade workflow builds on a burgeoning literature that examines the causal effects of texts on outcomes \citep{feder-etal-2022-causal}. \citet{fong2016sIBP} and \citet{fong2023causal} introduce a procedure for identifying the features of texts that drive responses, but their framework relies upon a more constrained topic model and was unable to infer the persuasiveness of new texts. \citet{egami_2022_CausalInf_SienceAdv} provide a general guide for causal inference with texts and outline a series of identification issues. \citet{PalmerStirling_2023_LLMs} experimentally demonstrate that LLMs can perform nearly as well as humans at producing persuasive arguments.

Other work has analyzed the linguistic characteristics of persuasive messages. \citet{feng-hirst-2011-classifying} categorize arguments into common schemes, while \citet{Tan2016_CMV_Reddit}, \citet{habernal-gurevych-2016-makes}, and \citet{gleize-etal-2019-convinced} examine successful arguments in online discussions, debate forums, and Wikipedia, investigating the predictive power of structural features.
\citet{wang-etal-2019-persuasion} explore personalized persuasion processes and \citet{wachsmuth-etal-2017-computational} propose a systematic taxonomy for argument quality. \citet{Zhang2020_CausalEffectsConversations} explore the causal effects of conversational tendencies. However, their work does not allow for domain-specific feature discovery and causal inference with these features. 
\citet{zhao-etal-2021-leveraging} model the relatedness among controversial topics using embedding-based methods based on individuals’ stances, integrating topic semantics from arguments and persuasion factors.
Currently, controllable argument generation relies upon previously identified features or domain expertise \citep{saha-srihari-2023-argu, schiller-etal-2021-aspect}. These procedures can augment steps 1 and 3 of the workflow.

\section{Conclusion}
\label{sec:Conclusion}
This paper introduces AutoPersuade---a new workflow for persuasion.  Our AutoPersuade approach curates arguments and collects responses to those messages, identifies the latent features that cause them to be more or less persuasive, infers the causal effects of those topics, and enables the selection of more persuasive messages from a new collection of candidate messages.  

 Each step of the workflow is modular and can be improved as new technologies become available. For example, better initial curation of arguments will make data collection more efficient; other interpretable models can be used to assess why some arguments are persuasive; and we can explore targeting of messages to particular people. New techniques could use the result of data and models to automatically generate more persuasive messages.

\section{Limitations and Ethics}
\label{sec:Limitations&Ethics}
Here we briefly overview the limitations and ethical considerations of our work.

\subsection{Limitations}
While our workflow is quite general, there are important limitations both to the general design and to our specific version of it. The main limitation of the framework is that it must be possible to collect a credible response variable from the relevant population to be persuaded. For example, in Section~\ref{sec:Vegansim}, we collect self-reported persuasion from Mechanical Turk workers; however, self-reports might substantially differ from induced behavioral change \citep{coppock2023persuasion} and Mechanical Turk workers may not be reflective of the population of interest. These messages must also have a sufficiently diverse set of argument features to be able to discover the ones which are most persuasive.
 This limitation is shared with other mechanisms of assessing messages like A/B tests.

 Our specific implementation also has important limitations. We are assuming that the document embeddings preserve the relevant information that allows for persuasion and that our topic model can pick it up. A more subtle concern is driven by the scaling invariance of the matrix factorization. The numerical value of the estimated effects is relative to the range of loadings for a given topic and thus is related to the distribution of that dimension in the training data. This means that while our estimates of the directional effect of topics are robust, the magnitude may not be. This is a problem without an obvious fix because there is no natural underlying scale to latent concepts.

In the applications reported above, we restrict ourselves to estimating the average persuasiveness of features---a limitation highlighted in our validation studies. A major opportunity moving forward would be to push past this general view and consider the effects of messages personalized to individual people based on some known covariates. This would naturally induce issues of power, but these might be addressable by moving beyond our static experimental design (where documents are assigned randomly) to an adaptive design which is optimized to find the most impactful message for each subpopulation \cite{offer-westort_coppock_green_2021_adaptiveDesign}. While these designs, which arise out of the literature on multi-armed bandits \citep{slivkins2019introduction} have been used for fixed message options, they would need to be modified to fit our setting.

\subsection{Ethics}
Persuasion is about convincing someone to do something they would otherwise not do. The ethical boundaries of persuasion are often viewed through the lens of what we are trying to persuade people to do. While we have chosen applications we see as ethically positive, these strategies can be used by other actors for applications we would not endorse---just like A/B testing. For example, \citet{mathur2023manipulative} demonstrate that politicians in the US use A/B testing to optimize messages in campaign emails. One could imagine a motivated actor using email opens as a response variable and learning even more effective techniques to induce responses from voters.  Whether those responses are ethically negative or positive depends on whether the email messages help voters realize their true preferences, or deceive voters into supporting a candidate they would not with better and more complete information.   

It is natural to worry that more effective persuasive tools will be used to persuade the public in a way that harms general welfare. These concerns arose as the radio reached most homes, they arose again when televisions became omnipresent, when the internet reached homes, when smartphones became widely available, when social media arrived on those smartphones, and similarly, they arise now with technologies like large language models. We think this history of concern over new technology is useful because it helps contextualize the current worries as an important and common reaction as new technologies are deployed in the public. Further, while we think our method is useful as an automated way to find persuasive tactics, it is important to note that persuasion itself has its limits.  It is exceedingly difficult to customize messages to audiences---say voters in an election---even with extensive marketing data \citep{hersh2015hacking}. 

% \section*{Acknowledgements}

% Bibliography entries for the entire Anthology, followed by custom entries
\bibliography{anthology,custom}
% Custom bibliography entries only
% \bibliography{custom}

\appendix
\clearpage
\section{Appendix}
\label{sec:appendix}

\subsection{Derivation of the Total Loss Function}
\label{sec:Appendix_Total_Loss}
\begin{align*}
    \mathcal{L} & = \alpha \mathcal{L}_A + (1-\alpha)\mathcal{L}_R\\
                & = \alpha \frac{1}{2}\left\|\mathbf{M}-\mathbf{W} \mathbf{B}\right\|_F^2  + (1-\alpha) \frac{1}{2}\left\|\mathbf{Y}-\mathbf{W} \boldsymbol{\gamma}  \right\|_2^2 \\ 
                & = \frac{\alpha}{2}\operatorname{tr}\left( (\mathbf{M}-\mathbf{W} \mathbf{B})^T(\mathbf{M}-\mathbf{W} \mathbf{B}) \right) \\
                & \quad +
                \frac{1- \alpha}{2}\left( (\mathbf{Y}-\mathbf{W} \boldsymbol{\gamma})^T(\mathbf{Y}-\mathbf{W} \boldsymbol{\gamma}) \right) \\
                & = \frac{\alpha}{2} \left[ 
                \| \mathbf{M} \|_F^2 -2 \operatorname{tr}\left( \mathbf{M}^T \mathbf{W}\mathbf{B} \right) +\| \mathbf{W}\mathbf{B} \|_F^2
                \right] \\
                & \quad +
                \frac{1-\alpha}{2} \left[ 
                \| \mathbf{Y} \|_2^2 - 2 \mathbf{Y}^T \mathbf{W} \boldsymbol{\gamma} + \| \mathbf{W} \boldsymbol{\gamma} \|_2^2 
                \right] \\
                % & = \frac{1}{2} \left[ 
                %  \alpha \| \mathbf{M} \|_F^2 + (1-\alpha) \| \mathbf{Y} \|_2^2  
                %  - 2 \bigg(\alpha \sum_{i=1}^n \sum_{j=1}^s M_{i,j} [\mathbf{W}\mathbf{B}]_{i,j} +(1-\alpha) \sum_{i=1}^n Y_i [\mathbf{W} \boldsymbol{\gamma}]_i\bigg) + \alpha \| \mathbf{W}\mathbf{B} \|_F^2 + (1-\alpha)  \| \mathbf{W} \boldsymbol{\gamma} \|_2^2 
                % \right] \\
                % & = \frac{1}{2} \left[ 
                % \left\| \left(\sqrt{\alpha}\mathbf{M} \big| \sqrt{1-\alpha} \mathbf{Y} \right) \right\|_F^2 \\
                % & \quad 
                % - 2 \left(  \sum_{i=1}^{n} \sum_{j=1}^{s+1} \left[ \left(\sqrt{\alpha}\mathbf{M} \big| \sqrt{1-\alpha} \mathbf{Y} \right) \right]_{i,j} \left[ \mathbf{W}\left( \sqrt{\alpha}\mathbf{B} \big| \sqrt{1-\alpha} \boldsymbol{\gamma} \right) \right]_{i,j} \right) \\
                % & \quad + 
                % \left\| \mathbf{W}\left( \sqrt{\alpha}\mathbf{B} \big| \sqrt{1-\alpha} \boldsymbol{\gamma} \right) \right\|_F^2 
                % \right] \\ 
                & = \frac{1}{2} \big\| \left(\sqrt{\alpha}\mathbf{M} \big| \sqrt{1-\alpha} \mathbf{Y} \right)  \\
                & \quad - \mathbf{W}\left( \sqrt{\alpha}\mathbf{B} \big| \sqrt{1-\alpha} \boldsymbol{\gamma} \right) \big\|_F^2 \\
                & = \frac{1}{2} \left\| \mathbf{X}  - \mathbf{W} \mathbf{H} \right\|_F^2 
\end{align*}
where $\mathbf{X}: = \left(\sqrt{\alpha}\mathbf{M} \big| \sqrt{1-\alpha} \mathbf{Y} \right)$ and $\mathbf{H} : = \left( \sqrt{\alpha}\mathbf{B} \big| \sqrt{1-\alpha} \boldsymbol{\gamma} \right)$.

\subsection{Semi-nonnegative Matrix Factorization}
\label{sec:appendix_SNMFupdates}

Following \citet{Ding2008_SemiNMF}, the closed form updating steps for the semi-nonnegative matrix factorization to minimize the total loss function of \eqref{eq:total_loss} are:

\begin{enumerate}
    \item[(S0)] Initialize $\mathbf{W} $. Do a $K$-means clustering. This gives cluster indicators $\mathbf{W} : \mathbf{W} _{i k}=1$ if $\mathbf{x}_i$ belongs to cluster $k$. Otherwise, $\mathbf{W} _{i k}=0$. Add a small constant to all elements of $\mathbf{W} $. Following \citet{Ding2008_SemiNMF}, we use 0.2. 
    \item[(S1)] Update $\mathbf{H}$ (while fixing $\mathbf{W} $ ) using the rule
    \begin{align*}
        \mathbf{H} & =[\mathbf X^T \mathbf{W} \left(\mathbf{W} ^T \mathbf{W} \right)^{-1}]^T\\ 
         & = (\mathbf{W}^T\mathbf{W})^{-1}\mathbf{W}^T\mathbf{X}
    \end{align*}
    
    Note $\mathbf{W} ^T \mathbf{W} $ is a $k \times k$ positive semidefinite matrix. The inversion of this small matrix is trivial. In most cases, $\mathbf{W} ^T \mathbf{W} $ is nonsingular. When $\mathbf{W} ^T \mathbf{W} $ is singular, we take the pseudoinverse.
    \item[(S2)] Update $\mathbf{W} $ (while fixing $\mathbf{H}$ ) using
    \begin{equation*}
          \mathbf{W}_{i k} \leftarrow \mathbf{W}_{i k} \sqrt{\frac{\left(\mathbf{X} \mathbf{H}^T\right)_{i k}^{+}+\left[\mathbf{W}\left(\mathbf{H} \mathbf{H}^T\right)^{-}\right]_{i k}}{\left(\mathbf{X} \mathbf{H}^T\right)_{i k}^{-}+\left[\mathbf{W}\left(\mathbf{H}  \mathbf{H}^T\right)^{+}\right]_{i k}}}  
    \end{equation*}
    where we separate the positive and negative parts of a matrix $M$ as  
    \begin{align*}
         M_{i k}^{+}& =\left(\left|M_{i k}\right|+M_{i k}\right) / 2, \\
         M_{i k}^{-}& =\left(\left|M_{i k}\right|-M_{i k}\right) / 2.
    \end{align*}
\end{enumerate}

Note that our implementation of this matrix factorization builds on \citet{PyMF} and  GitHub Copilot was used for the coding parts of this research.

\subsection{Data Collection and Preparation}

\subsubsection{Response Quality Control}
To ensure the quality of survey responses, we conducted a series of small pilot studies on Amazon Mechanical Turk (MTurk). Participants were initially selected based on their past acceptance rates and the number of completed tasks, but the results revealed mixed levels of response quality. However, we observed improved attention and response quality when we restricted the sample to English-speaking adults residing in the U.S. who held 'MTurk Master' status, a designation granted to users with a track record of consistently high-quality work.

For the main experiment, we collected 1,038 responses from MTurk Masters. Only two responses were rejected for incorrectly identifying the questionnaire's focus as arguments for political participation. All other respondents correctly recognized that the questionnaire concerned arguments for adopting a vegetarian/vegan diet or arguments against animal cruelty, resulting in 1,036 valid responses. This provided a total of 5,180 pairwise comparisons for evaluating the arguments.

We employed a pairwise forced-choice design, where participants compared two arguments at a time. This setup was chosen over ranking multiple arguments to reduce the cognitive load and memory demands on participants.

For the three validation studies, we similarly collected 198, 100, and 102 responses, respectively.

\subsubsection{Argument Selection for Validation Studies}
Following the curation of a set of \textit{proto-arguments}, we generated additional arguments using GPT-4, as outlined in Section \ref{sec:Validation_Studies}. This process resulted in the creation of 760 Synthesis Arguments and 60 Stronger Emphasis Arguments. For the first validation study, we selected 30 Synthesis Arguments and 15 Stronger Emphasis Arguments.

Two primary criteria guided the selection process. First, we filtered arguments that met the quantitative requirements: a higher predicted persuasiveness score compared to the original proto-arguments and the property that the two main latent topics of the proto-arguments are two topics with the highest loadings in the new arguments. From the pool of arguments that satisfied these criteria, we manually selected those with high predicted persuasiveness scores, ensuring diversity by excluding arguments that were overly similar. For instance, for each proto-argument, we generated three Stronger Emphasis Arguments; if two such arguments met the numeric thresholds but were highly similar, we only included one in the study.
After re-evaluating the argument filtering, we discovered that we included one Synthesis Argument that did not have a higher predicted persuasion score than both of its proto-arguments. However, excluding this argument and its pairwise comparisons from Validation Study 1 and 2 does not meaningfully affect the results.

For the third validation study, we employed a different argument generation strategy. We randomly selected arguments and prompted GPT-4 to rewrite them to either increase or decrease the presence of topic (2), \textit{inefficient use of resources}. We then filtered the revised arguments based on their inferred scores for topic (2), ensuring they reflected the intended changes. 
As with the first validation study, we ensured that the selected arguments were sufficiently distinct from one another, beyond their inferred topic loadings.

In all validation studies, the process of generating arguments by giving GPT-4 one or two initial inputs and specifying desired changes resulted in coherent arguments that were in line with the original argument collection. However, as expected, it was more challenging to increase the presence of topics that were already prominent in an argument, and it was similarly difficult to revise high-performing arguments to achieve even higher predicted persuasiveness scores. 
These relatively small margins in inferred topic loadings are less robust, and the filtering for Validation Study 1 is more affected by changing our topic inference method than the filtering for Validation Study 3, as discussed in Appendix \ref{Appendix:New_topic_inf}.

All the arguments utilized in validation studies are included in the Supplementary Materials.

\subsubsection{Alternative Embedding Models}
For our case study, we are utilizing using OpenAI’s "text-embedding-ada-002". However, we also tested and found almost identical predictive performance with both the new ‘small’ and ‘large’ embedding model 3 of OpenAI and the open-source SBERT paraphrase-MiniLM-L6-v2 model \citep{reimers-gurevych-2019-sentence}.  When inspecting a well-performing 10-topic model fit based on SBERT embeddings, we found that the identified topics roughly map pairwise to the 10 topics reported in this paper. Specifically, the correlation of topic loadings between these topic pairs across the 1308 original arguments ranged from 0.4 to 0.8, indicating that we can discover similar topics on embeddings derived from different models.

\subsubsection{Data Preparation and Processing}
We standardize our embedding representation and response variables to make the variance across the two data types approximately equal. This step ensures that neither the embedding nor the response variable mechanically dominates the loss function merely because the variance in one is much larger than the variance in the other.
In practice, we divided the 1536-dimensional embeddings used in the applications by 2 after we standardized them. 

Further, note that for any solution to the optimization problem, we can scale up $\widehat{\mathbf{W}}$ without affecting the result, as long as we scale down $\widehat{\mathbf{H}}$ accordingly and vice versa. To deal with this scale invariance problem, common across every matrix factorization task, we standardize the results.  We suggest dividing each column of $\widehat{\mathbf{W}}$ by its standard deviation, multiplying the rows of $\widehat{\mathbf{H}}$ with the corresponding standard deviations. 

\subsection{Additional Results - Persuasiveness of Arguments}

\begin{table}[ht]
    \centering
\scriptsize
    \begin{tabular}{lclc}
        \hline
\textbf{Dep. Variable:}    &  Pers. Score  & \textbf{  R-squared:         } &     0.259   \\
\textbf{Model:}            &      OLS      & \textbf{  Adj. R-squared:    } &     0.242   \\
\textbf{No. Observations:} &        505    & \textbf{  F-statistic:       } &     15.66   \\
\textbf{Covariance Type:}  &   nonrobust   & \textbf{  Prob (F-statistic):} &  2.31e-26   \\
        \hline
    \end{tabular}
    \vspace{5pt}  % Add vertical space between tables
\begin{tabular}{l D{.}{.}{4.4} D{.}{.}{3.3} D{.}{.}{3.3} D{.}{.}{3.3}}
& \multicolumn{1}{c}{\textbf{Coefficient}} 
& \multicolumn{1}{c}{\textbf{Std Err}} 
& \multicolumn{1}{c}{\textbf{t}} 
& \multicolumn{1}{c}{\textbf{P$> |$t$|$}} \\
\hline
\textbf{const} &     -0.0144  &        0.119     &    -0.121  &         0.904         \\
\textbf{(1)}   &      -0.0164  &        0.028     &    -0.585  &         0.559        \\
\textbf{(2)}   &       0.1319  &        0.028     &     4.717  &         0.000        \\
\textbf{(3)}   &       0.0233  &        0.032     &     0.737  &         0.462        \\
\textbf{(4)}   &      -0.1348  &        0.028     &    -4.836  &         0.000        \\
\textbf{(5)}   &      -0.0198  &        0.026     &    -0.763  &         0.446        \\
\textbf{(6)}   &       0.0569  &        0.028     &     2.044  &         0.042        \\
\textbf{(7)}   &      -0.1765  &        0.029     &    -6.115  &         0.000        \\
\textbf{(8)}   &       0.1041  &        0.029     &     3.642  &         0.000        \\
\textbf{(9)}   &      -0.0918  &        0.028     &    -3.242  &         0.001        \\
\textbf{(10)}  &       0.0393  &        0.029     &     1.341  &         0.180        \\
Arg. Length    &       0.0052  &        0.001     &     7.092  &         0.000        \\
    \hline
\end{tabular}
    \caption{Summary statistics of the causal effect estimation of the different topics discovered in the analysis of arguments for veganism including argument length (characters).}
\label{tab:Veganism_Estimation}
    \raggedright
\end{table}

\begin{figure}[ht]
    \centering
    \includegraphics[width=0.5\textwidth]{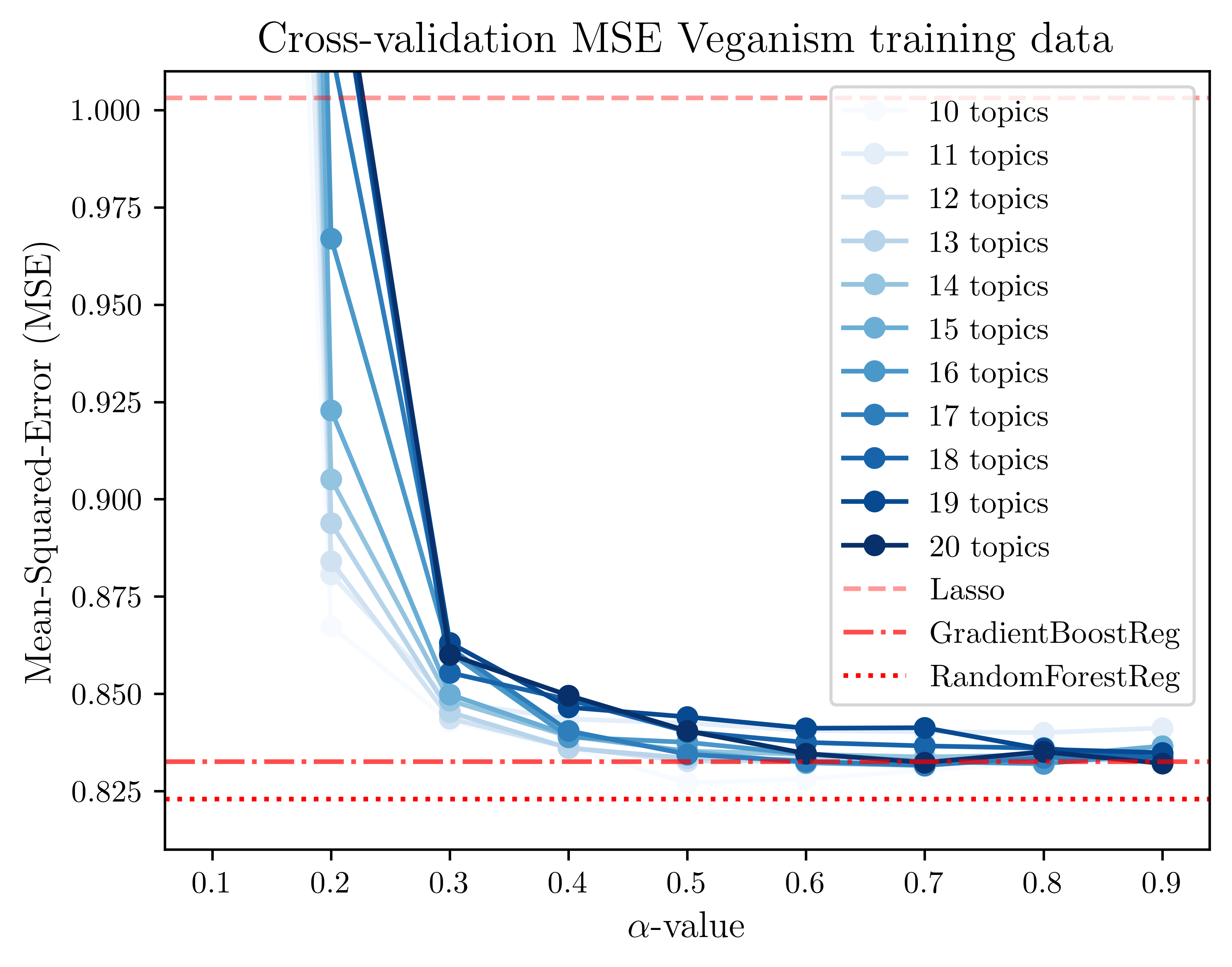}
    \caption{Out-of-sample predictive accuracy of SUN topic model for additional hyperparameter choices, as well as benchmark models on the training data. Results were calculated using 10-fold cross-validation.}
    \label{fig:Veganism_CV_MoreTopics}
\end{figure}

\begin{figure}[ht]
    \centering
    \includegraphics[width=0.5\textwidth]{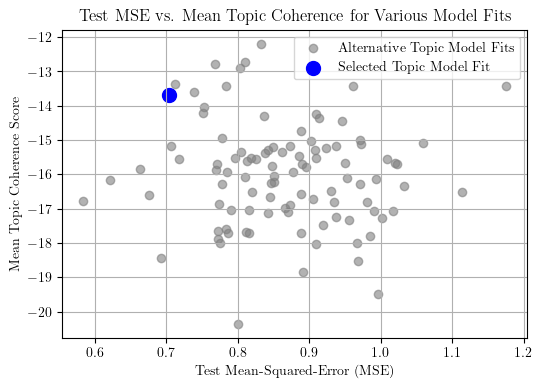}
    \caption{Topic coherence and out-of-sample predictive accuracy on 20\% holdout of the training data for parameter choices $\alpha=0.5$ and the number of topics $J=10$.}
    \label{fig:Veganism_Topic_Coherence}
\end{figure}

\subsubsection{Differentiation of Similar Topics}
The topic labels introduced in section \ref{sec:Veganism_Estimating_AMCE} encapsulate distinct themes, despite some apparent overlapping among Topics (4), (6), and (7). In particular, Topic (4), \emph{morals, ethics, and justifications}, emphasizes historical justifications for meat consumption, the role of societal norms, and the moral implications of human choices in eating meat or abusing animals. In contrast, Topic (7), \emph{animal rights and speciesism}, centers on the standing of animals as an oppressed group, discussing their rights, well-being, and the species-wide discrimination they face, often drawing parallels to other forms of historical oppression. Topic (6), \emph{individual contributions and responsibilities}, shifts the focus to the direct impact of personal actions, highlighting the cumulative effects of individual choices on alleviating suffering through conscious consumption.

\clearpage
\newpage
\section{Converging Topic Inference} 
\label{Appendix:New_topic_inf}
Following \citet{Ding2008_SemiNMF} and evaluations of the predictive performance in our cross-validation step, we use $100$ iterations of the semi-nonnegative matrix factorization updating steps when fitting our model. One step includes updating both $\mathbf{H}$ and $\mathbf{W}$. 
For our original results, reported above, we used the updating step for $\widehat{\mathbf{W}}$ while holding $\widehat{\mathbf{B}}$ fixed to minimize $\mathcal{L}_A$ to infer topic loadings for new documents based on a previously selected model fit. Matching our original topic fitting, we ran this updating step 100 times to infer topic loadings. 

However, the convex sub-problem of inferring $\widehat{\mathbf{W}}$ given $\widehat{\mathbf{B}}$ did not consistently converge with only 100 steps.
While this approach converges eventually, we also implemented a new extension to the SUN topic model where topic inference is done using the convex optimization solver CVXPY \citep{diamond2016cvxpy}. This allows for faster convergence.

In practice, this means that originally, when convergence was not met, the inferred topic loadings of a new document were marginally affected by the random initialization of its topic loadings and the set of documents (other rows) in $\widehat{\mathbf{W}}$ for which we simultaneously inferred topic loadings. 
While this does not have a meaningful effect on our causal estimates or predictive performance, the new inference method yields more robust results when inferring topic loadings. 

However, running $\widehat{\mathbf{W}}$ to convergence results in some documents with very high topic loadings across all latent topics, which reduces scarcity and complicates the interpretability of the inferred topic loadings. 
Identifying the ideal topic inference approach that balances robust results and the benefits of early stopping might be the subject of future research.

\subsection{Changes in Results}

While we find that the main results of our work do not change meaningfully, we include results corresponding to this updated topic inference approach in this section of the Appendix. 
As detailed in the following sections, we observe two main effects of the new inference method. First, the confidence intervals of our causal estimates are smaller, indicating a more precise inference of topic loadings. 
Second, some arguments selected for our validation studies no longer satisfy the filtering cutoffs based on inferred topic loadings. In particular, for Validation Study 1 \& 2, we selected and generated new arguments at the very tails of the topic loading distributions. The differences in topic loadings and predicted persuasiveness scores between original, proto-arguments, and newly generated arguments were often small. These small differences are affected by the new topic inference method, leading to different argument filtering results. 
However, our main results remain consistent when re-evaluating the validation studies, considering only arguments that passed the filtering using the newly inferred topic loadings.

\subsection{Cross-Validation}
\begin{figure}[ht]
    \centering
    \includegraphics[width=0.5\textwidth]{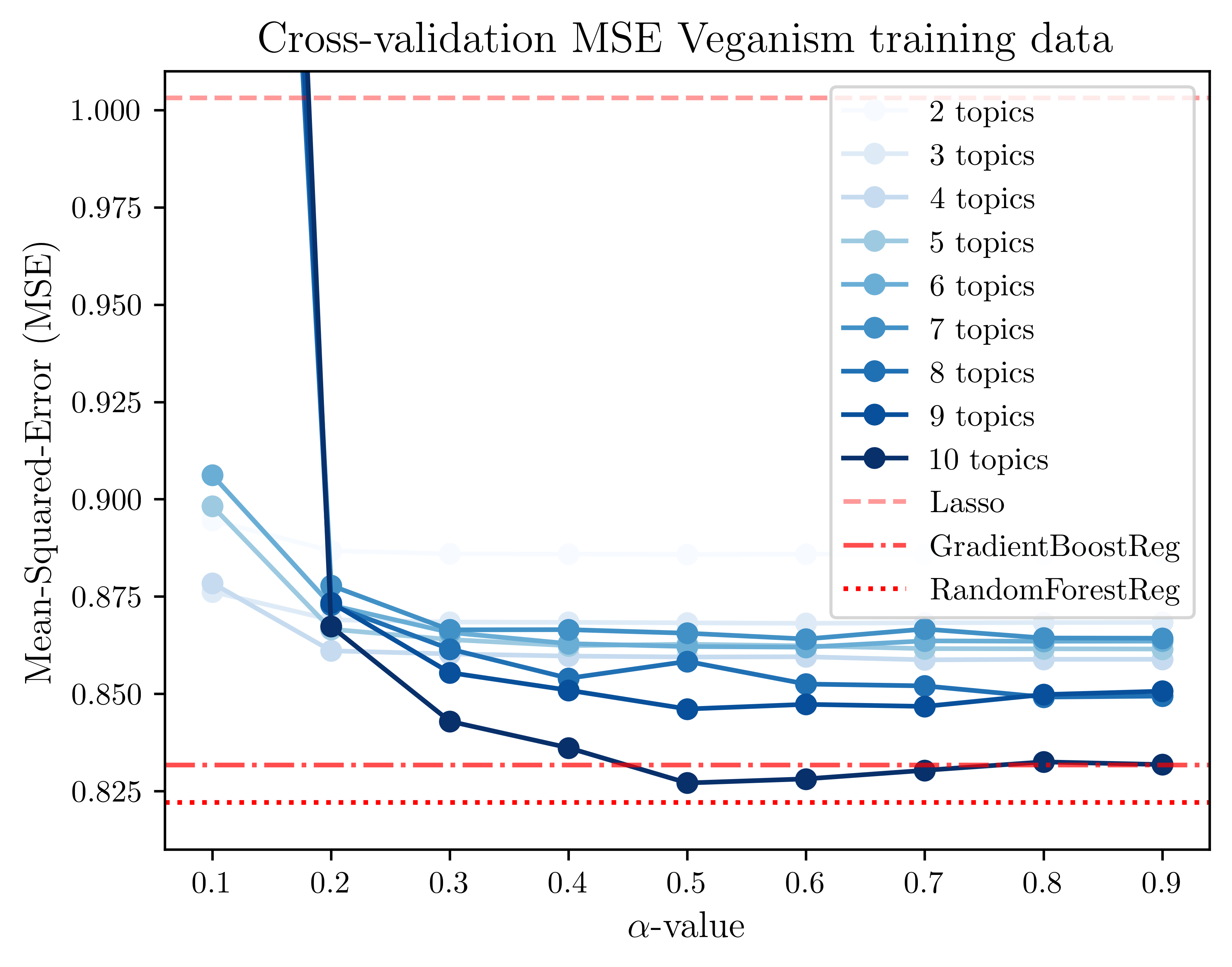}
    \caption{Out-of-sample predictive accuracy of SUN topic model for different parameter choices, as well as benchmark models on the training data. Results were calculated using 10-fold cross-validation and topic inference using CVXPY.}
    \label{fig:Veganism_CV_cvxpy}
\end{figure}

Figure \ref{fig:Veganism_CV_cvxpy} shows the cross-validation results using the CVXPY-based topic inference for the out-of-sample predictions. 
The results remain virtually unchanged and the combination of $J=10$ topics and $\alpha =0.5$ remains the best-performing hyperparameter choice.

\subsection{Causal Inference}
\begin{figure}[ht]
    \centering
    \includegraphics[width=0.5\textwidth]{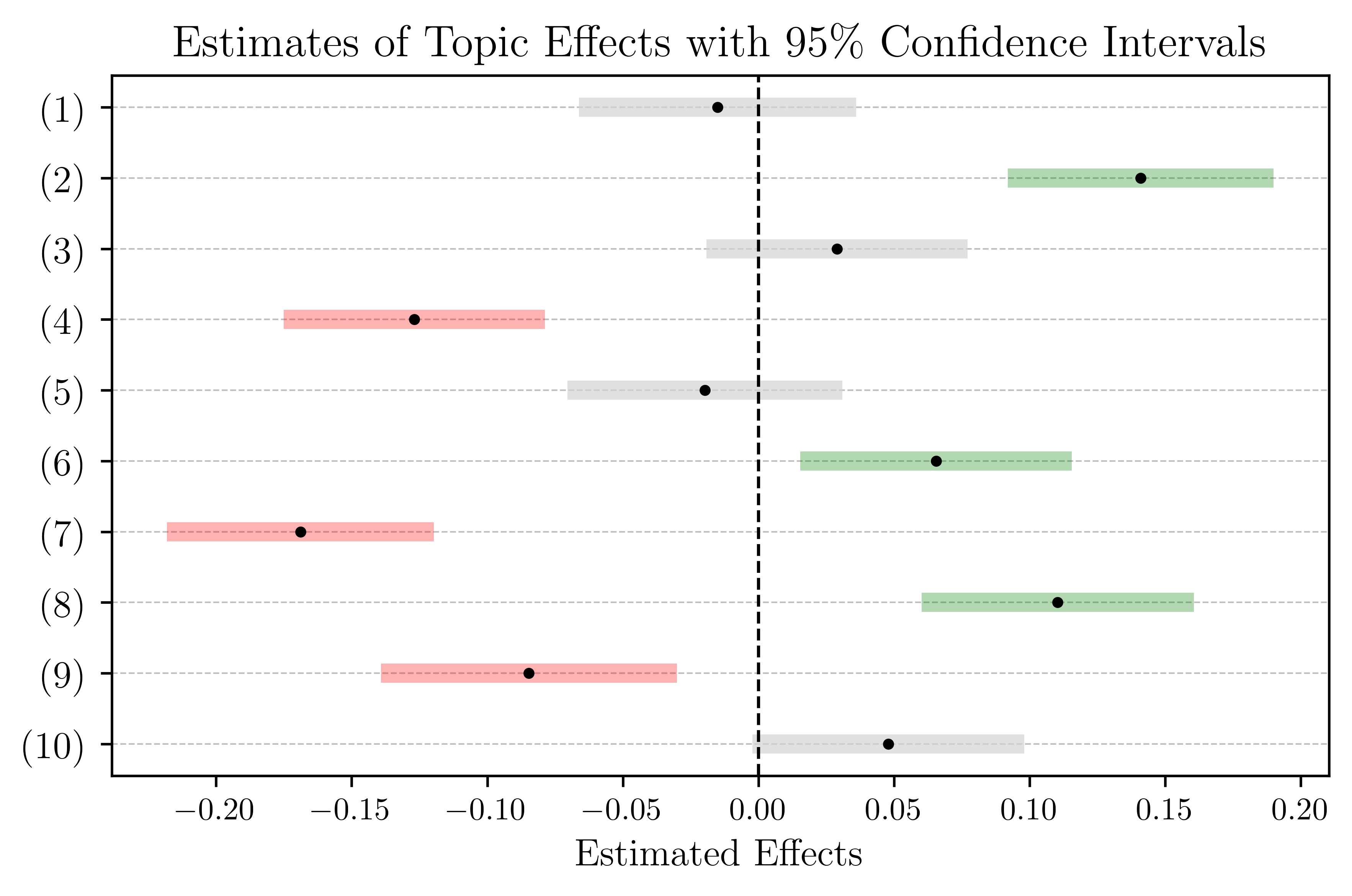}
    \caption{Estimated effects of discovered latent topics on the persuasiveness score of arguments for veganism, based on topic inferences using CVXPY. Refer to Table \ref{tab:Veganism_Estimation_cvxpy} for more details.}
    \label{fig:Veganism_Effects_cvxpy}
\end{figure}

Using the same topic model fit, we now infer the topic loadings on our estimation set using the new inference approach. 
As shown in Figure \ref{fig:Veganism_Effects_cvxpy} and Table \ref{tab:Veganism_Estimation_cvxpy}, our estimates remain mostly unchanged. However, we do observe smaller confidence intervals.

\begin{table}[ht]
    \centering
\scriptsize
    \begin{tabular}{lclc}
        \hline
\textbf{Dep. Variable:}    &  Pers. Score  & \textbf{  R-squared:         } &     0.259   \\
\textbf{Model:}            &      OLS      & \textbf{  Adj. R-squared:    } &     0.243   \\
\textbf{No. Observations:} &        505    & \textbf{  F-statistic:       } &     15.67   \\
\textbf{Covariance Type:}  &   nonrobust   & \textbf{  Prob (F-statistic):} &  2.25e-26   \\
        \hline
    \end{tabular}
    \vspace{5pt}  % Add vertical space between tables
\begin{tabular}{l D{.}{.}{4.4} D{.}{.}{3.3} D{.}{.}{3.3} D{.}{.}{3.3}}
& \multicolumn{1}{c}{\textbf{Coefficient}} 
& \multicolumn{1}{c}{\textbf{Std Err}} 
& \multicolumn{1}{c}{\textbf{t}} 
& \multicolumn{1}{c}{\textbf{P$> |$t$|$}} \\
\hline
\textbf{const}       &      -0.0612  &        0.035     &    -1.765  &         0.078    \\
\textbf{(1)}         &      -0.0151  &        0.026     &    -0.581  &         0.562    \\
\textbf{(2)}         &       0.1410  &        0.025     &     5.648  &         0.000    \\
\textbf{(3)}         &       0.0290  &        0.025     &     1.182  &         0.238    \\
\textbf{(4)}         &      -0.1269  &        0.025     &    -5.179  &         0.000    \\
\textbf{(5)}         &      -0.0197  &        0.026     &    -0.763  &         0.446    \\
\textbf{(6)}         &       0.0655  &        0.025     &     2.570  &         0.010    \\
\textbf{(7)}         &      -0.1689  &        0.025     &    -6.747  &         0.000    \\
\textbf{(8)}         &       0.1104  &        0.026     &     4.321  &         0.000    \\
\textbf{(9)}         &      -0.0847  &        0.028     &    -3.049  &         0.002    \\
\textbf{(10)}        &       0.0479  &        0.026     &     1.877  &         0.061       \\
\textbf{Arg. Length} &       0.0052  &        0.001     &     7.094  &         0.000       \\
    \hline
\end{tabular}
    \caption{Summary statistics of the causal effect estimation of the different topics discovered in the analysis of arguments for veganism and inferred using CVXPY.}
\label{tab:Veganism_Estimation_cvxpy}
    \raggedright
\end{table}

\subsection{Validation Studies}
While the previous results were only marginally affected by the new topic inference, the selection criteria for new arguments for our validation studies were more significantly impacted. 
As we are applying the relatively strict numerical filter for selecting arguments, we need to update our selected arguments for validation studies 1, 2, and 3.

In Validation Studies 1 and 2, we selected new arguments that had higher predicted persuasion scores and whose highest topic loadings corresponded to the targeted topics (two topics for synthesis arguments and one topic for stronger emphasis arguments).
When we use the newly derived topic loadings, only $17/30$ of the previously selected synthesis arguments and 10/15 of the stronger emphasis arguments meet these criteria. 

In Validation Study 3, we intervened to either increase or decrease the presence of topic (2), and then selected arguments that reflected the targeted change in their topic (2) loading.
Of the previously selected increased presence argument, we retain $67/80$, and of the previously selected decreased presence arguments, we retain $20/20.$

Only considering the outcomes of pairwise comparisons of arguments that we retain based on these new filtering results, we are left with $594/990$, $295/510$, and $434/500 $ pairwise comparisons for Validation Study 1, 2, and 3 respectively. 

We recalculate the win rates per argument group for Validation Studies 1, 2, and 3 as summarized in Table \ref{tab:Veganism_Validation_WinRates_New}.
While the lower number of comparisons leads to wider confidence intervals, there are no fundamental changes to the results of the validation studies. 
The main change is that Stronger Emphasis arguments are no longer the second best performing in terms of win rate in Validation Study 1.
Yet, all other relative performance rankings are preserved, and the main findings persist.

\begin{table}
    \centering
    \small
    \begin{tabular}{lrr}
    \toprule
     & Win Rate (\%) & 95\% CI \\
     \midrule
    \multicolumn{3}{l}{\textbf{Validation Study 1}} \\
    Argument Synthesis (SY) & \textbf{54.5} & [48.5, 60.2] \\
    Original (OG) & 45.2 & [40.5, 50.0] \\
    GPT-best (GPT) & \underline{53.2} & [46.4, 60.4] \\
    Stronger Emphasis (SE) & 51.6 & [44.4, 58.9] \\
    \midrule
    \multicolumn{3}{l}{\textbf{Validation Study 2}} \\
    Argument Synthesis (SY) & \underline{46.7} & [40.5, 53.3] \\
    Original (OG) & \textbf{54.2} & [48.8, 59.3] \\
    Stronger Emphasis (SE) & 41.8 & [29.1, 54.4] \\
    \midrule
    \multicolumn{3}{l}{\textbf{Validation Study 3}} \\
    Increased Topic (2) Args & 70.0 & [64.6, 74.8] \\
    Decreased Topic (2) Args & 30.6 & [22.3, 40.3] \\

    \bottomrule
    \end{tabular}
    \caption{Results of Validation Study 1, 2, and 3 when only considering arguments that pass the filtering using the CVXPY topic inference approach. For each study, win rates are calculated as the share of comparisons with arguments of other origins that are won. Confidence intervals are calculated based on 500 bootstraps of the individual comparison outcomes. The highest score for each study is bolded, the second-highest score is underlined.}
    \label{tab:Veganism_Validation_WinRates_New}
\end{table}

\newpage
\begin{landscape}
\begin{table}[ht]
\small
\centering
\renewcommand{\arraystretch}{1.4} % Increases row height for readability
\setlength{\tabcolsep}{6pt} % Adds padding between columns
\setlength{\arrayrulewidth}{0.3mm} % Thicker lines for structure

\begin{tabular}{|p{13.5cm}|*{10}{c|}} % Adjust column definitions
\toprule
\textbf{Argument} & \multicolumn{10}{c|}{\textbf{Topic Loadings}} \\ \cline{2-11} % Use \cline instead of \cmidrule to connect vertical lines
& \textbf{(1)} & \textbf{(2)} & \textbf{(3)} & \textbf{(4)} & \textbf{(5)} & \textbf{(6)} & \textbf{(7)} & \textbf{(8)} & \textbf{(9)} & \textbf{(10)} \\ \midrule
% \textbf{Argument} & \textbf{(1)} & \textbf{(2)} & \textbf{(3)} & \textbf{(4)} & \textbf{(5)} & \textbf{(6)} & \textbf{(7)} & \textbf{(8)} & \textbf{(9)} & \textbf{(10)} \\ \midrule
Many believe that most meat is sourced from humane, small farms, but the reality is factory farms are the major source. Their production processes are highly secretive, including the methods of slaughter. Even terms like 'free range' do not guarantee the absence of animal suffering during slaughtering. & \textbf{4.87} & 1.64 & 1.20 & 2.70 & 1.21 & 1.00 & 0.03 & 1.07 & 0.56 & 1.40 \\ \midrule
Slaughterhouses generally operate under a veil of secrecy and often deny external access. This lack of transparency raises questions about whether or not they adhere to ethically sound practices in the treatment of animals. & \textbf{4.71} & 0.33 & 0.84 & 2.52 & 0.90 & 0.71 & 0.57 & 1.57 & 0.61 & 1.63 \\ \midrule
Free range eggs may have a reputation for coming from chickens that live in idyllic settings. However, many hens remain living in confined, overcrowded sheds with limited access to daylight. Moreover, they often undergo distressing beak trimming measures. & \textbf{4.52} & 0.53 & 0.00 & 1.04 & 4.14 & 3.58 & 2.78 & 2.89 & 0.84 & 1.70 \\ \midrule
While it might not be universally accepted, it's a fact that battery farming practices can lead to hens being kept in smaller spaces that might seem inhumane. Regrettably, despite some regulatory efforts, these practices continue in many regions across the globe, including more developed regions like the USA and EU. & \textbf{4.43} & 0.94 & 0.00 & 0.50 & 3.20 & 3.20 & 3.02 & 2.16 & 1.77 & 2.93 \\ \midrule
One perspective is that animal farming might exploit laborers, possibly contributing to their physical and mental health stress. Critics argue that the sector evades workers' compensation and could potentially involve vulnerable individuals. & \textbf{4.39} & 1.13 & 1.04 & 0.13 & 0.87 & 0.27 & 1.87 & 1.92 & 2.50 & 2.80 \\ \midrule
\multicolumn{11}{|c|}{...} \\ \midrule 

Just as we humans value and deserve bodily autonomy, so do animals. Exploiting them for products like honey denies them their rights, erodes their freedom, and imposes our will on their natural existence. Embrace veganism to respect and uphold these rights. & \textbf{0.04} & 1.23 & 2.83 & 0.57 & 2.20 & 0.59 & 3.69 & 1.22 & 0.83 & 1.53 \\ \midrule
Choose veganism, contribute to water conservation. Producing cow's milk and beef necessitates more than triple the water used in making soya milk \& vegan burgers. Given escalating global water scarcity, adopting a vegan diet is a practical and impactful solution! & \textbf{0.04} & 3.83 & 2.09 & 0.72 & 2.22 & 1.32 & 0.65 & 0.93 & 3.34 & 1.44 \\ \midrule
Every single life, including those of animals, is precious and should be respected. We shouldn't sacrifice their existence to fulfill our dietary preferences. Upholding their right to life by adopting a vegan lifestyle is a compassionate choice that respects all beings. & \textbf{0.03} & 1.07 & 3.21 & 1.35 & 2.00 & 2.21 & 2.43 & 1.16 & 0.91 & 0.57 \\
\bottomrule
\end{tabular}
\caption{Overview of arguments with very high and very low loadings for Topic (1). \\This presents a preview of the documents that an analyst might inspect when developing the label for Topic (1). We place a strong emphasis on this manual inspection step when it comes to evaluating a topic model fit and deriving topic labels.}
\end{table}
\end{landscape}

\end{document}